\documentclass{article}
\usepackage{comment}
\usepackage{amssymb}
\usepackage{amsmath}
\usepackage{multirow}

\usepackage[preprint]{neurips_2020}
\usepackage{natbib}
\usepackage[utf8]{inputenc} 
\usepackage[T1]{fontenc}    
\usepackage{hyperref}       
\usepackage{url}            
\usepackage{booktabs}       
\usepackage{amsfonts}       
\usepackage{nicefrac}       
\usepackage{microtype}      
\usepackage{graphicx}
\usepackage{subcaption}
\usepackage[ruled,vlined]{algorithm2e}

\title{Physics-informed Tensor-train ConvLSTM for Volumetric Velocity Forecasting} 

\author{Yu Huang$^1$, Yufei Tang$^1$, Hanqi Zhuang$^{1}$\thanks{Corresponding author: Hanqi Zhuang} , James VanZwieten$^2$, Laurent Cherubin$^3$ \\
  $^1$ Department of Computer \& Electrical Engineering and Computer Science, Florida Atlantic University\\
  $^2$ Department of Civil, Environmental, and Geomatics Engineering, Florida Atlantic University \\
  $^3$ Harbor Branch Oceanographic Institute, Florida Atlantic University \\
  \texttt{\{yhwang2018,tangy,zhuang,jvanzwi,lcherubin\}@fau.edu} \\
}

\begin{document}
\maketitle

\begin{abstract}
According to the National Academies, a weekly forecast of velocity, vertical structure, and duration of the Loop Current (LC) and its eddies is critical for understanding the oceanography and ecosystem, and for mitigating outcomes of anthropogenic and natural disasters in the Gulf of Mexico (GoM). However, this forecast is a challenging problem since the LC behaviour is dominated by long-range spatial connections across multiple timescales. In this paper, we extend spatiotemporal predictive learning, showing its effectiveness beyond video prediction, to a 4D model, i.e., a novel Physics-informed Tensor-train ConvLSTM (PITT-ConvLSTM) for temporal sequences of 3D geospatial data forecasting. Specifically, we propose 1) a novel 4D higher-order recurrent neural network with empirical orthogonal function analysis to capture the hidden uncorrelated patterns of each hierarchy, 2) a convolutional tensor-train decomposition to capture higher-order space-time correlations, and 3) to incorporate prior physic knowledge that is provided from domain experts by informing the learning in latent space. The advantage of our proposed method is clear: constrained by physical laws, it simultaneously learns good representations for frame dependencies (both short-term and long-term high-level dependency) and inter-hierarchical relations within each time frame. Experiments on geospatial data collected from the GoM demonstrate that PITT-ConvLSTM outperforms the state-of-the-art methods in forecasting the volumetric velocity of the LC and its eddies for a period of over one week.

\end{abstract}

\section{Introduction}
As part of the North Atlantic western boundary current system, the Loop Current (LC) originates as the Yucatan Current, flows northward into the Gulf of Mexico (GoM), and veers east to exit through the Straits of Florida where it becomes the Florida Current, which then transitions into the Gulf Stream system. The LC dynamics is characterized by the shedding of clockwise (anticyclonic) rotating eddies that move westward further into the GoM. The LC and its eddies together are called Loop Current System (LCS). On April 20, 2010, the oil drilling rig Deepwater Horizon, operating in the Macondo Prospect in the GoM, exploded and sank resulting in 4 million barrels of oil gushing uncontrolled into the Gulf over an 87-day period, before it was finally capped on July 15, 2010 \citep{national2011deep}. This disaster, also known as GoM Oil Spill, threatened livelihoods, precious habitats, and even a unique way of life. Since the disaster, a large amount of research has improved scientific understanding and forecasting of the LC with aspirations for alleviating impacts on ecosystems and for future oil-spill prevention. The ability to predict LC evolution is critical and fundamental to almost all aspects of the GoM, including 1) anthropogenic and natural disaster response, 2) the prediction of short-term weather anomalies and, hurricane intensity and trajectories, 3) national security and safety, and 4) ecosystem services \citep{walker2009loop}. Due to a reinforcing interaction between seasonal hurricanes and the LCS, long-term prediction of the LCS states is of particular interest and becoming increasingly relevant for mitigating potential environmental and ecological damages. 

The National Academies of Science, Engineering and Medicine (NASEM) published a report \citep{national2018understanding} specifically calling for the development of models that are capable of forecasting a) current speed, vertical structure, and duration of the LC and its eddies a week ahead; b) LC propagation a month ahead; and c) eddy shedding events up to 13 weeks ahead. The aim of this research is to address the first capability request. In this paper, we develop a novel technique, named Physics-informed Tensor-train ConvLSTM (PITT-ConvLSTM), that enables 4D spatiotemporal LC prediction, specifically, prediction of a continuous sequence of 3D flow maps (see Section \ref{Section: dataset}). We address the problem of modeling the LCS based on convolutional LSTM networks (ConvLSTM) by incorporating prior domain knowledge described as nonlinear differential dynamic equations. Our PITT-ConvLSTM model can leverage required physics explicitly, and learn implicit patterns from data. Moreover, unlike most of the first-order ConvLSTM-based approaches, we propose a higher-order generalization to ConvLSTM with convolutional tensor-train decomposition \citep{su2020convolutional} to learn the long-term spatio-temporal structure in the LCS.

\section{Dataset}\label{Section: dataset}
\subsection{Data Description}
\begin{figure}[t]
  \centering
  \begin{subfigure}{0.46\textwidth}
  \includegraphics[width=\textwidth]{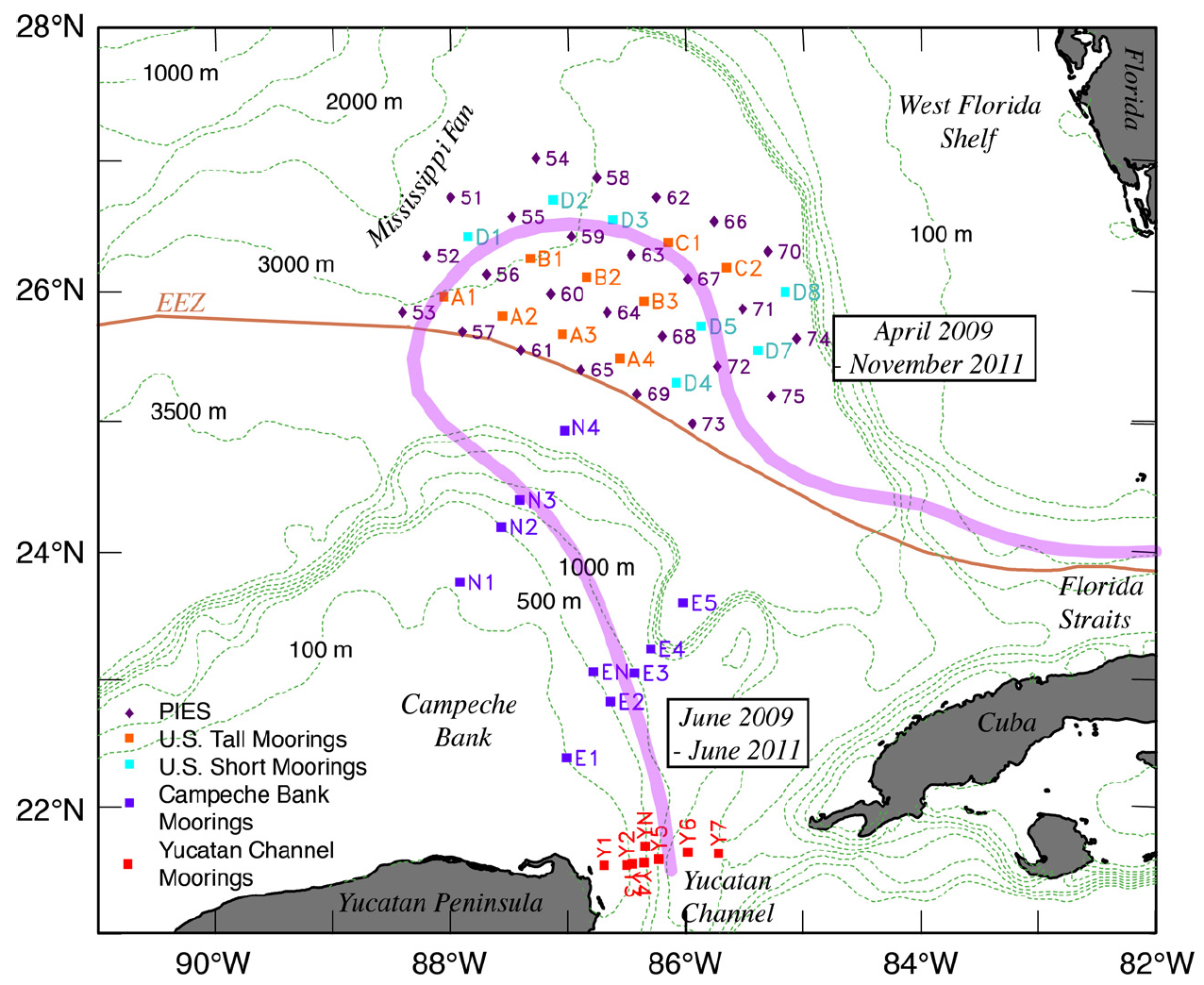}
  \caption{}
  \label{fig:data_location}
  \end{subfigure}
    \begin{subfigure}{0.53\textwidth}
  \includegraphics[width=\textwidth]{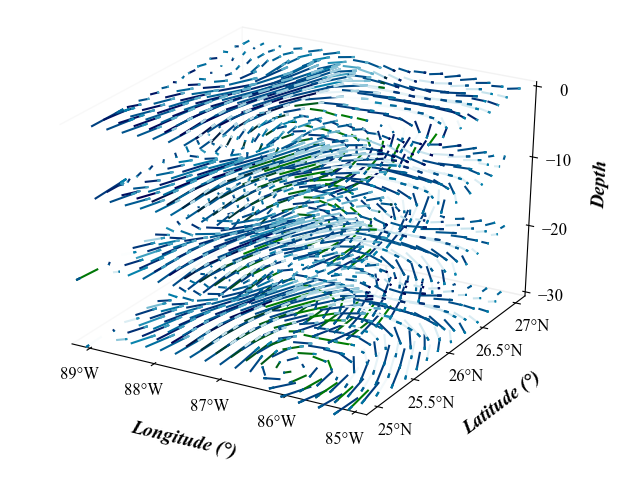}
  \caption{}
  \label{fig:data_clip}
  \end{subfigure}
  \caption{(a) Locations of moorings and PIES deployed in the U.S. and Mexican sectors in the eastern Gulf of Mexico \citep{hamilton2016loop}; (b) Visualization of a time slice $\mathcal{X}_t$ at 10 ($\times20m$) intervals.}
  \label{Fig.data}
\end{figure}

The dataset used in this study was produced by the observational program Dynamics of the Loop Current in US Waters Study \citep{hamilton2016loop}, which provides a 4D mapping of the current speed and density structure in the LC region from a high-density array of 149 moored instruments over a two-and-a-half-year period. The observational array (Figure \ref{Fig.data}(a)), from which the data were collected, covered the region where the LC extended northward and, more importantly, where eddy shedding events occurred most often. This sensor array consisted of 25 inverted echo sounders with pressure gauges, 9 full-depth tall moorings with temperature, conductivity and velocity measurements, and 7 near bottom current meter moorings deployed under the LC region. From the measurements gathered by these sensors, the geostrophic velocities of the region were computed (Figure \ref{Fig.data}(b)). The complete procedure for producing these mapped velocity fields is described in \citep{hamilton2016loop} and \citep{donohue2016loop}.

The dataset contains velocity data gathered from June 2009 to June 2011 over the region from $89^o W$ to $85^o W$, and $25^o N$ to $27^o N$ with 30–50 $km$ horizontal resolution. Sampling frequency from the multiple sensors varied from minutes to hours. In this study, we utilize a time series which was processed with a fourth order Butterworth filter and sub-sampled at 12-hour intervals. Consequently, the dataset contains a total of 1810 records, covering 905 days. At each time step $t$, the data $\mathcal{X}_t$ were formatted as $\mathcal{X}_t \subseteq \mathbb{R}^{D\times H\times W\times C}$, where each dimension represents $Depth$ (30), $Longitude$ (29), $Latitude$ (36) and $Channel$ (2), respectively. $Channel$ stands for the geostrophic velocity vector $(u,v)$. The first 80\% (June 2009 to December 2010) of the dataset was reserved for training and the remaining 20\% (January 2011 to June 2011) for prediction and validation.

\subsection{Empirical Orthogonal Function} 
Empirical Orthogonal Function (EOF) analysis has been extensively used in the oceanic and atmospheric sciences. In addition to its ability to decompose a time series into its temporal and spatial components, EOF can drastically reduce the dimension while preserving data integrity. Considering the nonlinear dynamics and high dimension characteristics of oceanic phenomena, in this work, the EOF analysis is employed to represent spatial patterns ($EOFs$) and temporal components (Principal Components, $PCs$) of the LCS in the GoM. Specifically, EOF analysis was conducted to extract the $PCs$ of zonal and meridional velocity of ocean current. The $PCs$ may contain otherwise hidden and medium-term uncorrelated patterns \citep{navarra2010guide}. Further, $PCs$ are used to train the PITT-ConvLSTM prediction model, with details given in Section \ref{Section: methods}. 

Singular Value Decomposition (SVD) is a general decomposition to determine both the $EOFs$ and $PCs$ simultaneously. For the convenience of ocean modeling, the ocean is sliced into layers in the depth direction. Here, SVD is applied to each depth slice $\mathcal{S}$ to obtain two unitary orthogonal matrices ($U$ and $V$) and one diagonal eigenvalue matrix $\Sigma$. Consequently, $U\Sigma$ represents the temporal $PCs$ and $V^T$ the spatial $EOFs$. Thus, $\mathcal{S}$ can be represented as follows,
\begin{equation}\label{Eqn:eof}
\mathcal{S} = U\Sigma V^T \leftrightarrow \underbrace{\begin{bmatrix}s_{1,1} & \cdots & s_{1,m}\\ \vdots &\ddots & \vdots \\ s_{n,1} & \cdots & s_{n,m}\end{bmatrix}}_{\mathcal{S}} = \underbrace{\begin{bmatrix}p_{1,1} & \cdots & p_{1,m}\\ \vdots &\ddots & \vdots \\ p_{n,1} & \cdots & p_{n,m}\end{bmatrix}}_{PCs} \underbrace{\begin{bmatrix}e_{1,1} & \cdots & e_{1,m}\\ \vdots &\ddots & \vdots \\ e_{m,1} & \cdots & e_{m,m}\end{bmatrix}}_{EOFs}
\end{equation}

After the EOF analysis, we concatenate the flattened $PCs$ along the depth dimension. In this way, the raw data $\mathcal{X}\subseteq \mathbb{R}^{T\times D\times H\times W\times C}$ is compressed to ${\mathcal{X}}'\subseteq \mathbb{R}^{T\times D\times P\times C}$, where $P$ is the number of selected elements in $PCs$. The 3D volume sequence prediction $\{\mathcal{X}^{(t)}\}_{t=1}^T$, in consequence, is compressed to a 2D matrix sequence prediction $\{\mathcal{X}'^{(t)}\}_{t=1}^T$. Each row of the 2D matrix $\mathcal{X}'^{(t)}(i,:)$ (flattened PCs) represents a depth slice of the original 3D volume $\mathcal{X}^{(t)}(i,:,:)$. The model in Section \ref{Section: methods} will be trained to predict future $PCs$, which could also be used to reconstruct predicted 3D volume by simply multiplied by the known $EOFs$ (constant) as shown in Equation (\ref{Eqn:eof}).

\section{Method}\label{Section: methods}
\subsection{Method Overview}
\begin{figure}
  \centering
  \includegraphics[width=\textwidth]{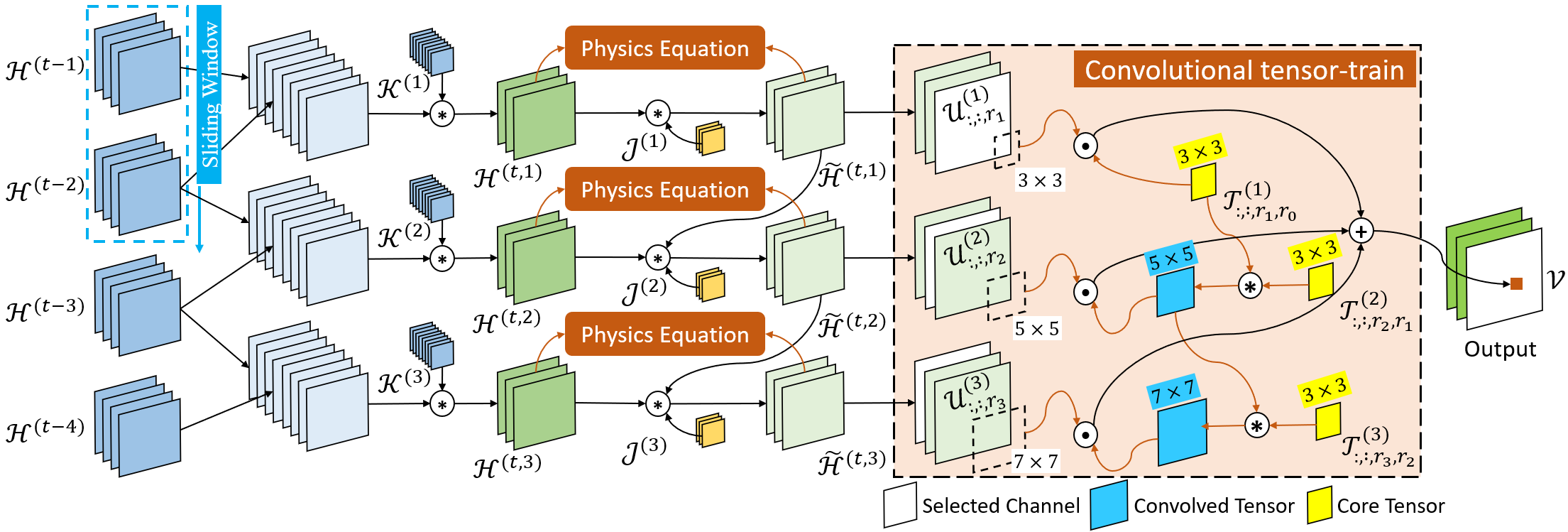}
  \caption{Schematic of $\varphi\{\mathcal{T}, \mathcal{H}\}$ within the PITT-ConvLSTM cell.}
  \label{Fig.frame}
\end{figure}

Convolutional LSTM network (ConvLSTM), a basic building block for sequence-to-sequence prediction, demonstrated strong performance in video forecasting. In ConvLSTM, the spatial information is encoded explicitly as tensors in the LSTM cells, where each cell is a first-order Markovian model (i.e. the hidden state is updated based on its adjacent step). Since ConvLSTM is successful in modeling complex behaviors and extracting abstract features through real-world data \citep{xingjian2015convolutional}, it is natural to explore how such a predictive model can be used to solve practical problems in physics or engineering domains with higher-order dynamics. In this paper, we propose Physics-informed Tensor-train ConvLSTM (PITT-ConvLSTM), a ConvLSTM network under physics constraints. The ConvLSTM is integrated with convolutional tensor-train to model higher-order spatio-temporal correlations explicitly, with its hidden states incorporated with prior physical knowledge. The proposed model is illustrated in Figure \ref{Fig.frame} and Figure \ref{Fig.architecture}. 

\subsection{ConvLSTM with Convolutional Tensor-train Decomposition}\label{Section: CTTD}
ConvLSTM, proposed by \citep{xingjian2015convolutional} as a convolutional counterpart of conventional FC-LSTM, introduces convolution tensor operation into input-to-state and state-to-state transitions within each recurrent cell. In ConvLSTM, all features are encoded as third-order tensors with dimensions ($ Height \times Width \times Channels$). At each time step $t$, a ConvLSTM cell updates its hidden states $\mathcal{H}^{(t)}$ based on the previous $\mathcal{H}^{(t-1)}$ and the current input $\mathcal{X}^{(t)}$:
\begin{equation}
\label{Eqn:lstm}
    \left [ \mathcal{I}^{(t)},\mathcal{F}^{(t)},\tilde{\mathcal{C}}^{(t)},\mathcal{O}^{(t)} \right ] = \sigma\left ( \mathcal{W}_{input} \ast \mathcal{X}^{(t)} + \mathcal{T}_{hidden} \ast \mathcal{H}^{(t-1)} \right )
\end{equation}
where $\sigma(\cdot)$ applies sigmoid on the input gate $\mathcal{I}^{(t)}$, forget gate $\mathcal{F}^{(t)}$, and output gate $\mathcal{O}^{(t)}$, and $tanh(\cdot)$ on memory cell $\tilde{\mathcal{C}}^{(t)}$. The $\mathcal{I}^{(t)},\mathcal{F}^{(t)},\mathcal{O}^{(t)}, \tilde{\mathcal{C}}^{(t)} \subseteq \mathbb{R}^{H\times W\times C} $, and the parameters are characterized by two 4-$th$ order tensors $\mathcal{W}_{input}\subseteq \mathbb{R}^{K\times K\times S\times 4C}$ and $\mathcal{T}_{hidden}\subseteq \mathbb{R}^{K\times K\times C\times 4C}$, where $K$ is the kernel size, and $S$ and $C$ are the numbers of input channels.

To capture multi-steps spatio-temporal correlations in ConvLSTM, we introduce a higher-order recurrent unit, where the hidden state $\mathcal{H}^{(t)}$ is updated based on the current input $\mathcal{X}^{(t)}$ and its $n$ previous steps $\left\{ \mathcal{H}^{(t-l)} \right\}^n_{l=1}$ with an $m$-order convolutional tensor-train decomposition $\varphi(\left\{\mathcal{T}^{(l)} \right\}^m_{l=1},\cdot)$ as follows:
\begin{equation}
    \left\{\tilde{\mathcal{H}}^{(t,o)}\right\} = f\left ( \mathcal{J}^{(o)}, \mathcal{K}^{(o)}, \left\{\mathcal{H}^{(t-l)} \right\}^n_{l=1}\right), \forall o \subseteq[m]
    \label{Eqn:map}
\end{equation}
\begin{equation}
\label{Eqn:cttdlstm}
    \left [ \mathcal{I}^{(t)},\mathcal{F}^{(t)},\tilde{\mathcal{C}}^{(t)},\mathcal{O}^{(t)} \right ] = \sigma\left ( \mathcal{W}_{input} \ast \mathcal{X}^{(t)} + \varphi\left\{ \left\{\mathcal{T}^{(o)} \right\}^m_{o=1}, \left\{\tilde{\mathcal{H}}^{(t, o)} \right\}^m_{o=1} \right\} \right )
\end{equation}
where the $m$-order convolutional tensor-train $\varphi(\left\{\mathcal{T}^{(l)} \right\}^m_{l=1},\cdot)$ are parameterized by $m$ core tensors $\left\{\mathcal{T}^{(l)} \right\}^m_{l=1}$. Considering the consistency constraint, the previous $n$ steps are mapped into $m$ intermediate tensors ${\tilde{\mathcal{H}}^{(t,o)}}$ by $f(\mathcal{J},\mathcal{K},\cdot)$ in Equation (\ref{Eqn:map}) at first, where $\mathcal{K}$ and $\mathcal{J}$ are 3D convolutional kernels. Note that $f(\mathcal{J},\mathcal{K},\cdot)$ is a mapping function with dynamic physics constraints, which will be discussed in Section \ref{sec:physics}.

The convolutional tensor-train decomposition, first proposed by \citep{su2020convolutional}, is a counterpart of tensor-train decomposition (TTD) which aims to represent a higher-order tensor $\mathcal{T}\subseteq \mathbb{R}^{I_1\times \cdots \times I_m}$ in a set of smaller and lower-order core tensors $\{\mathcal{T}^{(l)}\}^m_{l=1}$ with $\mathcal{T}^{(l)}\subseteq \mathbb{R}^{I_l\times R_l \times R_{l-1}}$. The ranks $\{R^{(l)}\}^m_{l=1}$ here control the number of parameters in the tensor-train format. In this way, the original $\mathcal{T}$ of size $\prod_{l=1}^{m}I_l$ is compressed to $\sum_{l=1}^{m}I_lR_{l-1}R_l$, i.e. the complexity only grows linearly with the order $m$ (assuming $R_l$’s are constants). Similar to TTD, $\varphi$ is designed to significantly reduce both parameters and operations of higher-order spatio-temporal recurrent models by factorizing a large convolutional kernel into a chain of smaller kernels. The details of convolutional tensor-train decomposition $\mathcal{T} = \mathcal{CTTD}\left\{\mathcal{T}^{(l)} \right\}^m_{l=1}$ is formulated as:
\begin{equation}\label{eqn:cttd}
    \mathcal{T}_{:, r_1,r_{m+1}} \triangleq \sum_{r_2=1}^{R_2}\cdots\sum_{r_m=1}^{R_m}\mathcal{T}_{:,r_1,r_2}\ast\cdots\ast\mathcal{T}_{:,r_m,r_{m+1}}^m
\end{equation}
and the convolutional tensor-train for spatial-temporal modeling can be equivalently stated as:
\begin{equation}\label{eqn:cttdmodeling}
    \mathcal{V}_{:,r_{l+1}}^{l+1} = \sum_{r_l=1}^{R_l} \mathcal{T}_{:, r_l,r_{l+1}}^l\ast(\mathcal{V}_{:,r_l}^{l}+\mathcal{U}_{:,r_l}^{l})
\end{equation}
where $\mathcal{U}$ is the input feature corresponding to $\tilde{\mathcal{H}}$. $\left\{\mathcal{V}^l\right\}_{l=1}^m$ are intermediate results, in which $\mathcal{V}^l\subseteq\mathbb{R}^{H\times W\times R_l}$ for $l>1$, and $\mathcal{V}^1\subseteq\mathbb{R}^{H\times W\times R_m}$ is initialized as all zeros and final prediction is returned as $\mathcal{V}=\mathcal{V}^{m+1}$. The exact mathematical proof is included in Appendix A.

\subsection{Physics Constraints}\label{sec:physics}
In this section, we provide how the law of physics that describes the physical processes (e.g., the LCS) can be incorporated into the deep learning framework. Physical processes are often modeled by a set of nonlinear partial differential equations (PDEs), which describe how a physical quantity is changed in a given region over time. Such dynamic equations are usually written as a relation between time derivatives and spatial derivatives:
\begin{equation}
    \sum_{i=1}^{P}b_i(u,t)\frac{\partial^i u}{\partial t^i}= \sum_{i=1}^{Q}c_i(u,x)\frac{\partial^i u}{\partial x^i}
\end{equation}
where $u$ is a physical quantity, say, velocity, and $x$ is its spatial coordinate. Furthermore, coefficients $b_i$, $c_i$ are functions of $u$ and $x$. Finally, $P$ and $Q$ denote the highest order of time derivatives and spatial derivatives, respectively.  

There are multiple ways to incorporate the physical laws into neural networks. In this paper, inspired by \citep{seo2019differentiable}, we model the law of physics by replacing the updating functions in neural networks with corresponding operators. Considering the characteristics of ocean current, we adopt two dynamic equations, i.e. the diffusion equation and the wave equation, shown in Table \ref{table:pde}. The diffusion equation describes the behavior of the continuous physical quantities (macroscopic behavior of many micro-particles in Brownian motion) resulting from the random movement. The wave equation is a second-order PDE for the description of waves (e.g. water, sound, or seismic waves).

\begin{table}[h]
\caption{Ocean Current related Dynamic Equations}
\label{table:pde}
\centering
\small
\begin{tabular}{@{}ll@{}}
\toprule
Updating Function & Dynamic Equation \\ \midrule
${v}'_i = v_i+\alpha(\bigtriangleup v)_i$ &  Diffusion: $\dot{u}=\alpha\bigtriangledown^2u$\\
${v}''_i = 2{v}'_i-v_i+c^2(\bigtriangleup {v}')_i$ &  Wave: $\ddot{u}=c^2\bigtriangledown^2u$\\ \bottomrule
\end{tabular}
\end{table}

In Equation (\ref{Eqn:map}), a sliding window strategy is adopted. As shown in Figure \ref{Fig.frame}, a sliding subset of $\{\mathcal{H}^l\}$ are concatenated and then transformed into $m$ (i.e. the number of core tensors in $\varphi$) intermediate hidden tensor $\{\mathcal{H}^{(t,o)}\}$. The concrete process is as follows:
\begin{equation}
    \mathcal{H}^{(t,o)}=\mathcal{K}^o\ast\left[ \mathcal{H}^{(t-n+m-l)}; \cdots; \mathcal{H}^{(t-l)}\right]
\end{equation}
where $n$ is the sliding window size. The $\{\mathcal{H}\}^n_{l=1}$ are first concatenated into $m$ tensors $\{\mathcal{H}_{cat}\}^m_{l=1}$ ($\mathcal{H}_{cat}\subseteq \mathbb{R}^{H\times W\times (n-m+1)\times C}$) along the time axis, which are thereafter mapped to $\mathcal{H}^{(t,o)}\subseteq \mathbb{R}^{H\times W\times R}$ by the 3D convolutional kernel $\mathcal{K}^o\subseteq \mathbb{R}^{K\times K\times (n-m+1)\times R}$.

Then, we update the intermediate hidden tensor $\{\mathcal{H}^{(t,o)}\}$ to $\{\tilde{\mathcal{H}}^{(t,o)}\}$ with physics constraints (see Pseudocode in Appendix B). Finally, the sequentially updated intermediate hidden tensor are transformed to the output space by convolutional tensor-train decomposition described in Section \ref{Section: CTTD}. It is notable that the law of physics is not directly constrained to the raw observations, but rather to the latent representations (i.e. the hidden states of ConvLSTM in this work). This is a desired configuration considering the difficulty in identifying which observations are following the law of physics explicitly and how much in accordance. Consequently, instead of individually applying the equation to each observation, we found that it is more efficient to introduce the constraints on the latent representations.

As illustrated in Figure \ref{Fig.frame}, $\mathcal{J}$ is the learn-able parameter vector while the orange blocks (``Physics Equation'') are objective functions related to physics constraints. First, we define physics constraints between the previous and updated states based on the known/assumed knowledge as:
\begin{equation}
    \mathcal{L}_{dp}^{(t,o)}=\emph{g}\left(\mathcal{H}^{(t,o)}, \tilde{\mathcal{H}}^{(t,o)}\right)
\end{equation}
where $\emph{g}$ is case-specific. In particular, if we are aware/assume that the observations should have a diffusive property, the diffusion equation can be used as the physics-informed constraint as:
\begin{equation}\label{Eqn: gdif}
    \emph{g}_{dif} = \left \| {v}'-v-\alpha\bigtriangledown^2v \right \|^2
\end{equation}
where $v$ (or ${v}'$) is the corresponding element of $\mathcal{H}$ (or $\tilde{\mathcal{H}}$).

The physics-informed objective function of the total sequence is defined as:
\begin{equation}
    \mathcal{L}_{dp}=\sum_{t}^{T}\sum_{o=1}^{m}\emph{g}\left(\mathcal{H}^{(t,o)}, \tilde{\mathcal{H}}^{(t,o)}\right)
\end{equation}
and the overall objective function is the sum of the supervised loss and physical loss: 
\begin{equation}\label{Eqn:loss}
    \mathcal{L} = \mathcal{L}_1 + \mathcal{L}_2 + \lambda\mathcal{L}_{dp}
\end{equation}
where $\mathcal{L}_1$ is the mean absolute error, $\mathcal{L}_2$ is the mean squared error, and $\lambda$ controls the importance of physics term.

\section{Results and Discussion}

\begin{figure}
  \centering
  \includegraphics[width=\textwidth]{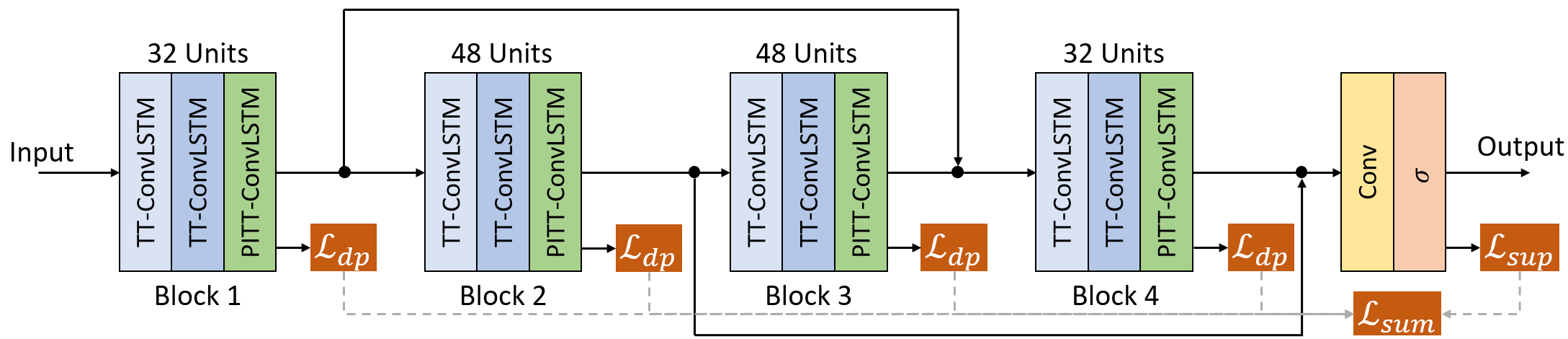}
  \caption{Illustration of the recurrent network architecture of 12-layer.}
  \label{Fig.architecture}
\end{figure}

\textbf{Model Details}. All experiments, shown in Figure \ref{Fig.architecture}, use a stack of 4 blocks (3 stacked layers of TT-ConvLSTM or PITT-ConvLSTM per block) and two skip connections added between the 1st and 3rd, 2rd and 4th blocks that perform concatenation over channels \citep{byeon2018contextvp}. The channels are set to 32 for the 1st and 4th blocks and 48 for the middle blocks. A convolutional layer is applied on top of all the recurrent layers to compute the predicted frames. The PITT-ConvLSTM is with order 3, step 3, rank 8. All models are trained with ADAM optimizer \citep{kingma2014adam} with objective function in Equation (\ref{Eqn:loss}). Once the model does not improve in 20 epochs (in terms of validation loss), scheduled sampling \citep{bengio2015scheduled} is then activated to ease the training with linearly decreased sampling ratio from 1 to 0. Learning rate decay is further activated if the loss does not drop in 20 epochs, and the rate is decreased exponentially by 0.98 every 5 epochs. The initial learning rate is set to $10^{-4}$. All models are trained to predict 10 frames given 10 input frames.

\textbf{Multi-Steps Prediction}. Two metrics are adopted to evaluate the performance and provide frame-wise quantitative comparisons. Specifically, the mean squared error (MSE) is for element-wise difference, while the structural similarity index measure (SSIM) \citep{wang2004image}, ranges between -1 and 1, is for perceptual similarity.

\begin{table}[h]
\caption{Comparisons of 10, 20 and 30 time-steps prediction (equal to 5, 10 and 15 days ahead forecast), where lower MSE (in $10^{-3}$) or higher SSIM indicates better model performance.}
\centering
\small
\begin{tabular}{@{}cccccccc@{}}
\toprule
\multirow{2}{*}{Methods} & \multicolumn{2}{c}{$10\rightarrow 10$} & \multicolumn{2}{c}{$10\rightarrow 20$} & \multicolumn{2}{c}{$10\rightarrow 30$} & 
\multirow{2}{*}{Parms}\\ \cmidrule(l){2-7} 
& MSE & SSIM & MSE & SSIM & MSE & SSIM \\ \midrule
ConvLSTM \citep{xingjian2015convolutional} & 9.257 & 0.623 & 18.584 & 0.465 & 26.725 & 0.374 & 4.94m \\
PredRNN \citep{wang2017predrnn} & 8.752 & 0.648 & 19.229 & 0.490 & 27.512 & 0.351 & 8.84m\\
TT-ConvLSTM \citep{su2020convolutional} & 8.575 & 0.656 & 17.445 & 0.498 & 23.802 & 0.420 & \textbf{0.97}m\\ \midrule
PITT-ConvLSTM (diffusion) & 7.429 & 0.686 & 15.835 & 0.527 & 22.426 & 0.439 & 0.99m\\
PITT-ConvLSTM (wave) & \textbf{6.971} & \textbf{0.687} & \textbf{14.097} & \textbf{0.543} & \textbf{19.638} & \textbf{0.463} & 0.99m\\\bottomrule
\end{tabular}
\label{table:result}
\end{table}

\begin{figure}[h]
\begin{subfigure}{0.5\textwidth}
\includegraphics[width=0.92\linewidth]{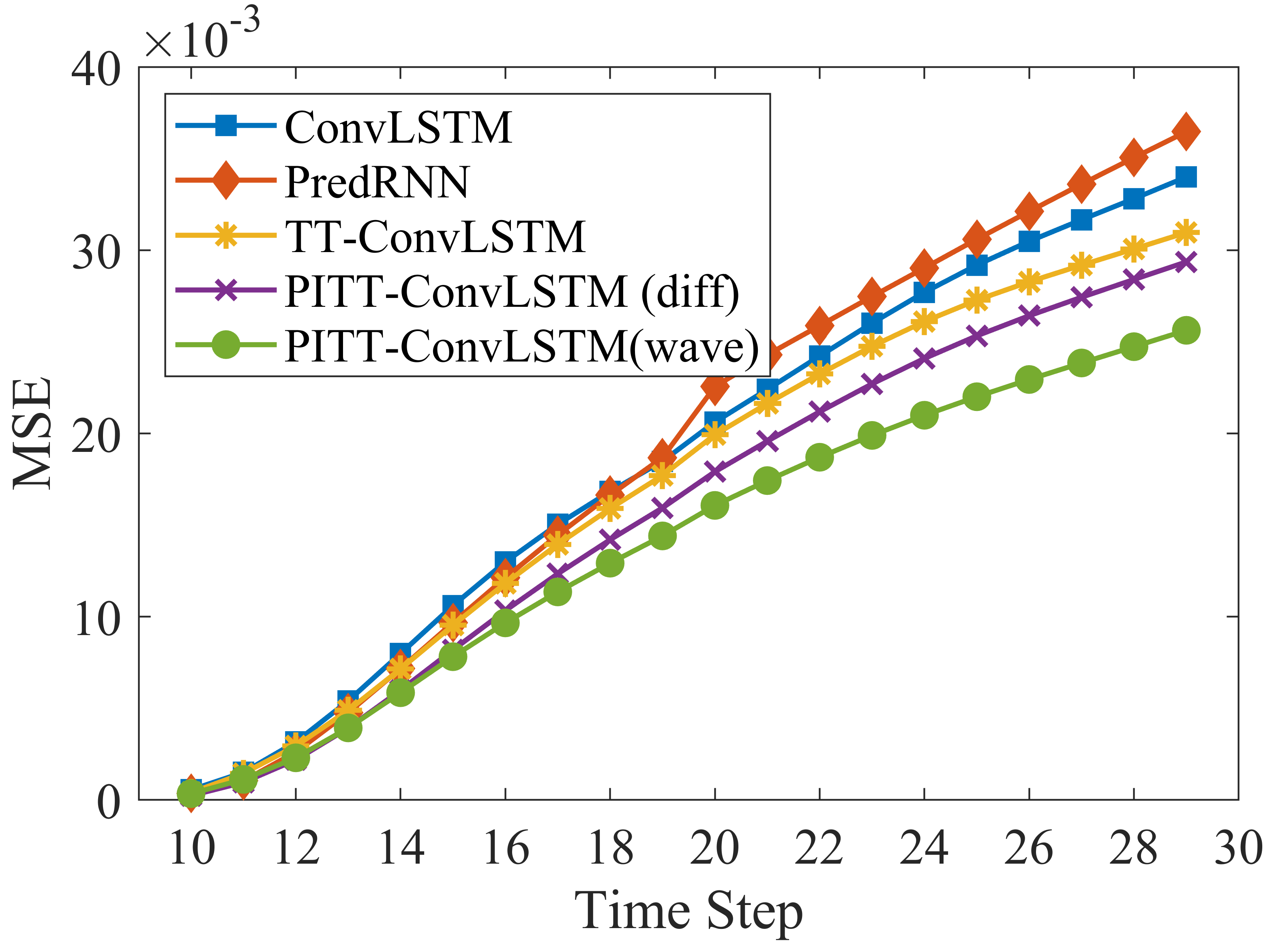} 
\label{fig:subim1}
\end{subfigure}
\begin{subfigure}{0.5\textwidth}
\includegraphics[width=0.92\linewidth]{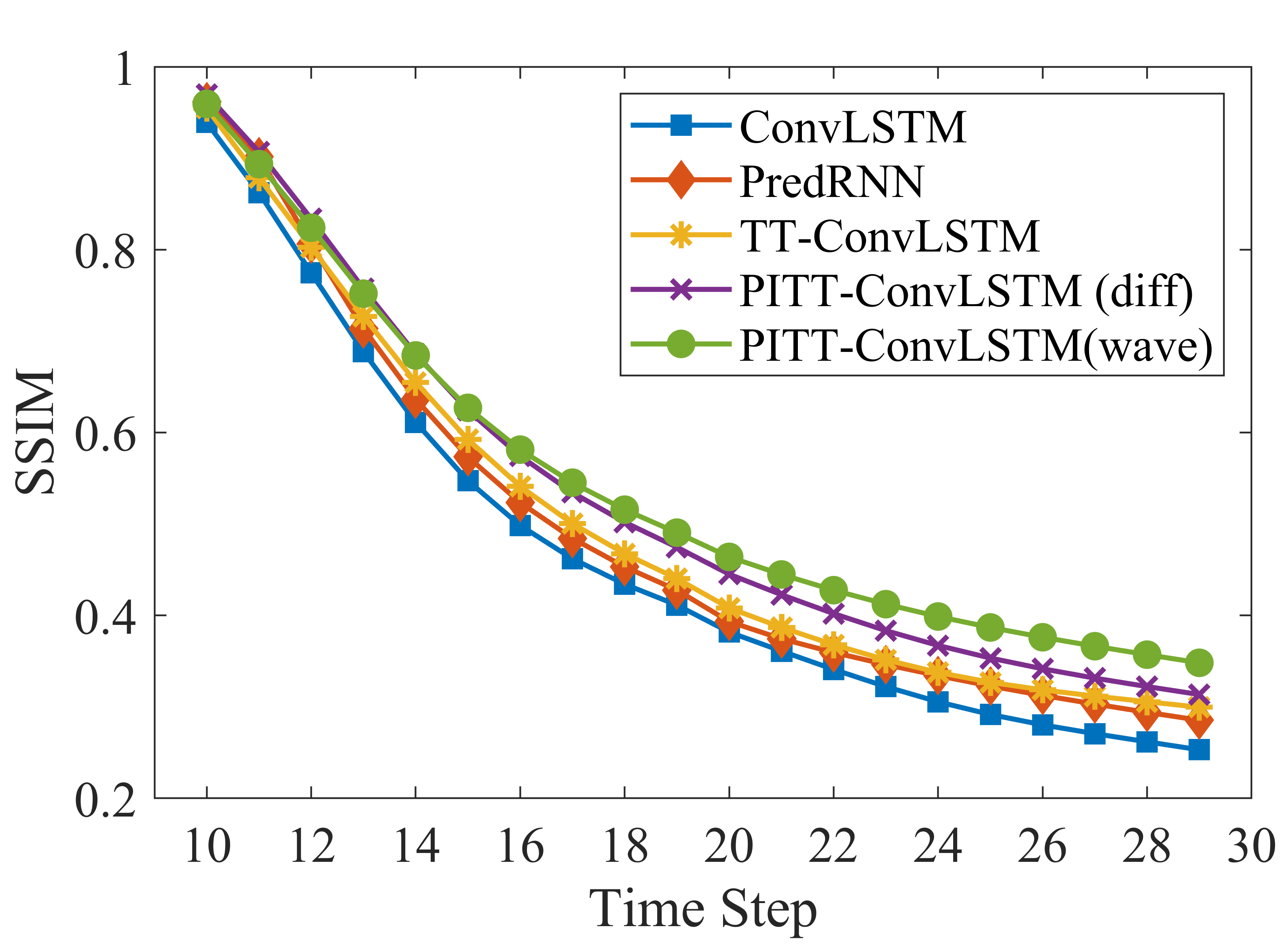}
\label{fig:subim2}
\end{subfigure}
\caption{Frame-wise comparison in MSE and SSIM.} 
\label{Fig:result}
\end{figure}

Table \ref{table:result} reports the average statistics for 10, 20, and 30 frames prediction. Figure \ref{Fig:result} shows comparisons of per-frame statistics. The ConvLSTM generates blurred future frames (lowest SSIM), since it fails to memorize the detailed spatial representations. By contrast, results by our proposed PITT-ConvLSTM in both metrics remain stable over time, with only a comparatively slow and reasonable decline. Also, it is notable that the methods using convolutional tensor-train decomposition, TT-ConvLSTM and PITT-ConvLSTM, achieve better performance with fewer parameters. 

Further, we explore how much the physics constraints helped improve LC prediction. Among the CCTD based models, PITT-ConvLSTM provides the least MSEs and highest SSIM. It proves that it is valid reasoning to incorporate physical rules in latent representing learning, since knowing physic rule constrained neighboring information is helpful to infer its own states. Specifically, the wave-informed PITT-ConvLSTM consistently outperforms all baseline models and shows superior predicting power both spatially and temporally. The LC prediction is partially visualized in Figure \ref{fig:preds}. We provide additional visual comparison among ConvLSTM, PredRNN, and TT-ConvLSTM in Appendix C.

\section{Related Work}
\subsection{Loop Current Prediction}
Model-based LC prediction primarily uses finite-differences or finite-element techniques to discretize PDEs in numerical models. The classical Princeton Regional Ocean Forecast System \citep{oey2005exercise} used data assimilation techniques to constrain their model solution with remote and in situ observations. Later, a local ensemble transform Kalman filter was applied to Princeton Ocean Model \citep{xu2013hindcasts} to estimate sea surface heights and LC temperatures. An ensemble based optimal interpolation technique \citep{counillon2009high} was further used to assimilate altimetry in the Hybrid Coordinate Ocean Model to predict LC eddy shedding. In the GoM 3-D Operational Ocean Forecast System Pilot Prediction Project \citep{mooers2012final}, several mesoscale eddy-resolving baroclinic ocean circulation numerical models were evaluated. With the popularization of machine learning, \citep{zeng2015predictability} proposed a novel model based on artificial neural network and EOF analysis for the LCS evolution prediction. Recently, a Divide-and-Conquer prediction model based on the LSTM network was developed to forecast the evolution of the LC and its eddy shedding \citep{wang2019medium}. It was capable of predicting the LC evolution for a period of 9 weeks, and the timing and general location of eddy Darwin's shedding event 12 weeks in advance, and eddy Cameron's detachment and reattachment 8 weeks in advance.
\begin{figure}
\centering
\begin{subfigure}[b]{\textwidth}
\centering
\includegraphics[width=\textwidth]{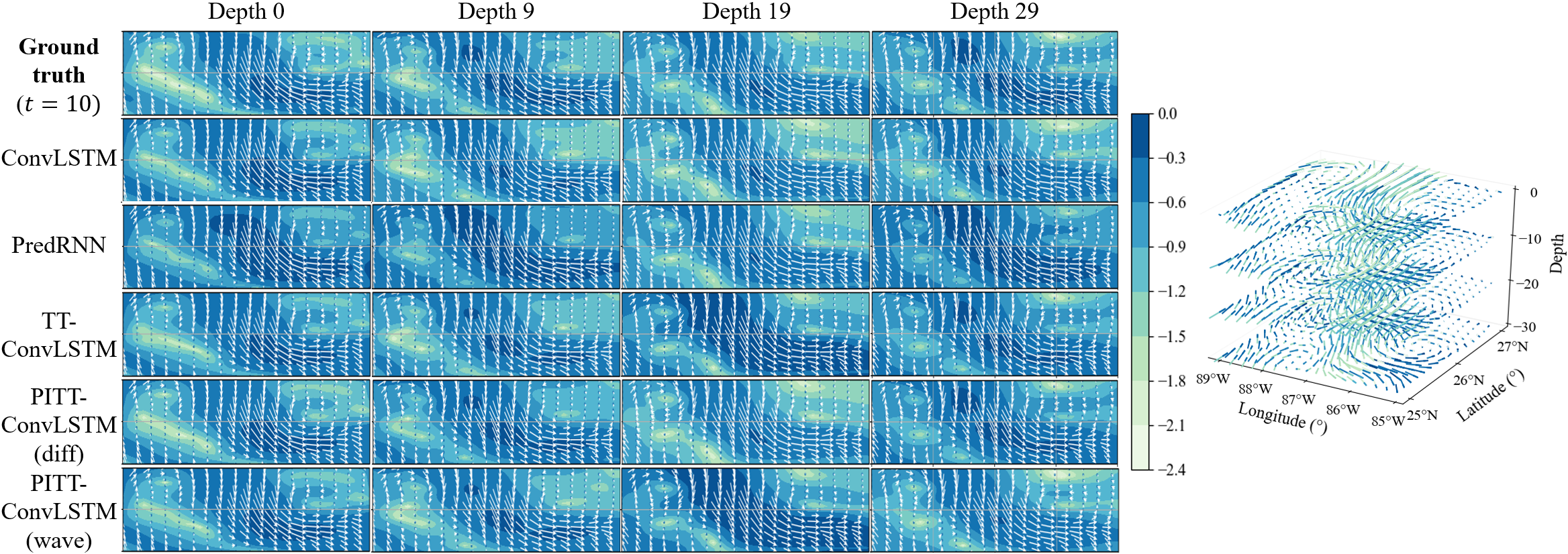}
\end{subfigure}
\begin{subfigure}[b]{\textwidth}  
\centering 
\includegraphics[width=\textwidth]{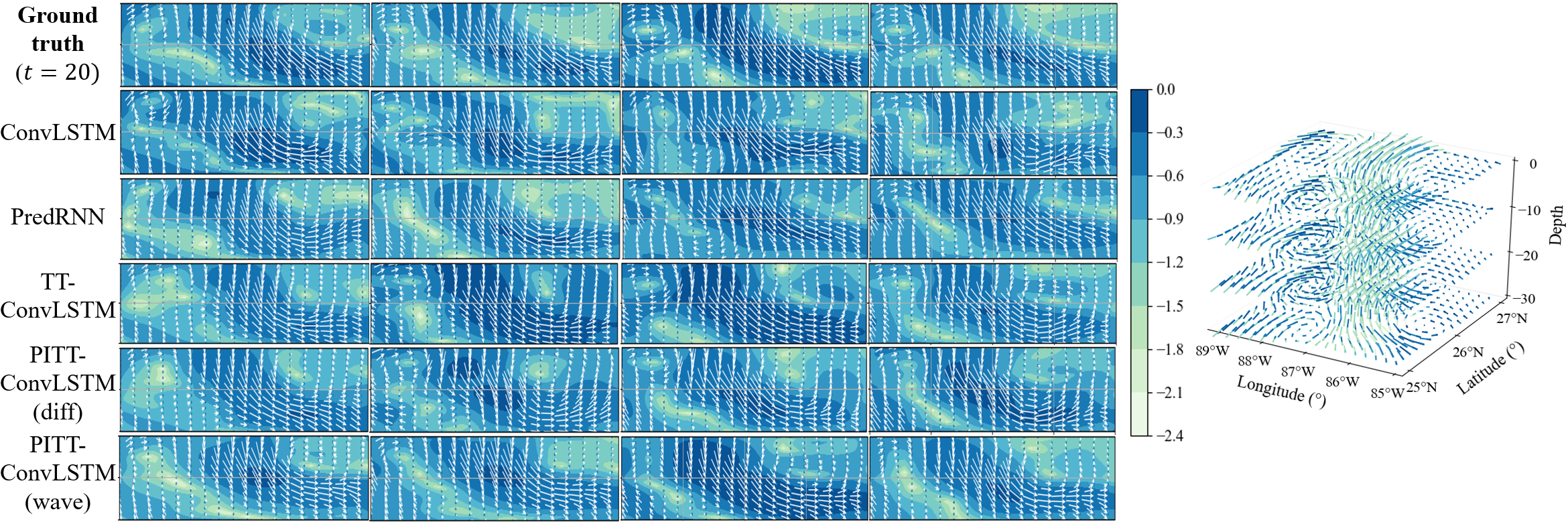}
\end{subfigure}
\caption{The visualization of LCS prediction. Every 10 ($\times12h$) steps ahead prediction with every 10 ($\times20m$) depth slices for each model are illustrated. The colormap indicates log-based LC velocity magnitude, i.e., $\lg \sqrt{u^2+v^2}$.} 
\label{fig:preds}
\end{figure}
\subsection{Recurrent Neural Network with Tensor-train Decomposition}
TTD, as a principled approach that factorizes the model parameters into smaller tensors \citep{oseledets2011tensor} for addressing the curse of dimensionality, is especially beneficial in multi-relational data analysis, and has been widely used in machine learning, including CNNs \citep{kolbeinsson2019robust,su2018tensorized,su2020convolutional}, RNNs \citep{yang2017tensor,su2020convolutional} and transformers \citep{ma2019tensorized}. A multiplicative RNNs with factorized weights tensor was proposed for inputs-states interactions \citep{sutskever2011generating}. Further, \citep{yang2017tensor} factorized the input-to-hidden weights within each cell by TTD, and showed improvement in video classification. Different from the first-order RNNs mentioned above, higher-order RNNs \citep{soltani2016higher} require excessively more parameters since they introduce connections cross multiple previous time steps for better long-term dynamics learning. This was improved by TT-RNNs \citep{yu2018long}, whose higher-order structures within each cell were compressed by TTD while improving the model performance in video classification. Recently, \citep{su2020convolutional} introduced a novel convolutional TTD and constructed Convolutional Tensor-Train LSTM to capture higher-order space-time correlations.

\subsection{Physics-informed Neural Networks}
Physics is one of the fundamental pillars explaining how nature behaves. Concerning that it is unable to describe all rules governing real-world data, machine learning is desired to bridge the known physics and observations. \citep{raissi2017physics,raissi2018hidden,raissi2018deep} demonstrate that neural networks are capable of finding solutions to PDEs, which enables us to obtain fully differentiable physics-informed models with respect to all input coordinates and free parameters. \citep{de2019deep} demonstrates how fluid physics could be incorporated for forecasting sea surface temperature, and such method not only captures the dominant physics but also infers unknown patterns by CNNs. Further, growing physics-informed models \citep{sanchez2018graph,kipf2018neural} were developed, assuming that neural networks can learn complex dynamic interactions and simulate unseen dynamics based on a current state. Unlike those works that implicitly extracts latent patterns from data only, the physics-informed Graph networks proposed in \citep{seo2019differentiable} allow incorporating known physics and simultaneously extracting latent patterns in data which is unable to be captured by existing knowledge.

\section{Conclusion}
In this paper, we proposed a physics-informed tensor-train ConvLSTM, a higher-order RNN, that is capable of effectively capturing long-term spatio-temporal correlations in temporal sequence of volumetric data. Within the PITT-ConvLSTM cell, a large convolutional kernel was factorized into a set of smaller core tensors through convolutional tensor-train decomposition. In the learnt latent space, physical domain knowledge in the form of PDEs over time and space were incorporated to facilitate the learning. The performance of our proposed PITT-ConvLSTM was demonstrated on volumetric velocity forecasting of LCS in the GoM, and the results verified that our model can produce superior results compared to state-of-the-art models. In a future study, we aim to design an effective prediction model that can capture simultaneously both fast changing local patterns and slow varying global trends.

\section*{Broader Impact}
Understanding the dynamics of the loop current system is fundamental to understanding the Gulf of Mexico’s full oceanographic system, and vice versa. Hurricane intensity, offshore safety, oil spill response, the fishing industry, and the Gulf Coast economy are all affected by the position, strength, and structure of the LC and associated eddies. The LC’s position varies greatly from its retracted state in the Yucatan Channel, directly east of the Florida Straits, to its extended state into the far northern and western Gulf. Why and when the LC suddenly intrudes north has not been able to be predicted with sufficient skill. The Physics-informed tensor-train ConvLSTM method proposed in this paper achieves significant improvements in forecasting the LC speed, vertical structure, and duration out to a forecast period of a few days to over 1 week. This research also recommends a strategy for addressing the key gap between physical process and deep learning in general understanding of LCS processes, in order to instigate a significant improvement in both short-term and long-range predictions of the LCS, which will increase overall understanding of GoM circulation and to promote safe oil and gas operations and disaster response in the GoM. 



\bibliographystyle{plainnat}
\bibliography{ref.bib}

\end{document}